\pdfoutput=1

\documentclass[11pt]{article}

\usepackage[]{ACL2023}
\usepackage{tcolorbox}
\usepackage{graphicx}    
\usepackage{subcaption}  

\usepackage{times}
\usepackage{fontawesome}
\usepackage{amsmath}
\usepackage{colortbl}
\usepackage{latexsym}
\usepackage{amsfonts}
\usepackage[T1]{fontenc}
\usepackage{booktabs}
 \usepackage{multirow}

\usepackage[utf8]{inputenc}

\usepackage{microtype}

\usepackage{inconsolata}
\usepackage{graphicx}

\title{A Context-Aware Contrastive Learning Framework for \\Hateful Meme Detection and Segmentation}

\author{Xuanyu Su,  
  Yansong Li, 
  Diana Inkpen \\
  University of Ottawa\\
  Ottawa, ON, Canada, K1N 6N5\\
  \texttt{\{xsu072,yli627,diana.inkpen\}@uottawa.ca} \\\And
  Nathalie Japkowicz\\
  American University\\
  Washington, DC, USA, 20016-8058\\
  \texttt{japkowic@american.edu} \\}

\begin{document}
\maketitle
\begin{abstract}
Amidst the rise of Large Multimodal Models (LMMs) and their widespread application in generating and interpreting complex content, the risk of propagating biased and harmful memes remains significant. Current safety measures often fail to detect subtly integrated hateful content within ``Confounder Memes''. To address this, we introduce \textsc{HateSieve}, a new framework 
designed to enhance the detection and segmentation of hateful elements in memes. \textsc{HateSieve} features a novel Contrastive Meme Generator that creates semantically correlated memes, a customized triplet dataset for contrastive learning, and an Image-Text Alignment module that produces context-aware embeddings for accurate meme segmentation. Empirical experiments show that \textsc{HateSieve} not only surpasses existing LMMs in performance with fewer trainable parameters but also offers a robust mechanism for precisely identifying and isolating hateful content.
\textcolor{red}{Caution: Contains academic discussions of hate speech; viewer discretion advised.} 

\end{abstract}

\section{Introduction}
\begin{figure}[t]
    \centering
    \includegraphics[width=0.49\textwidth]{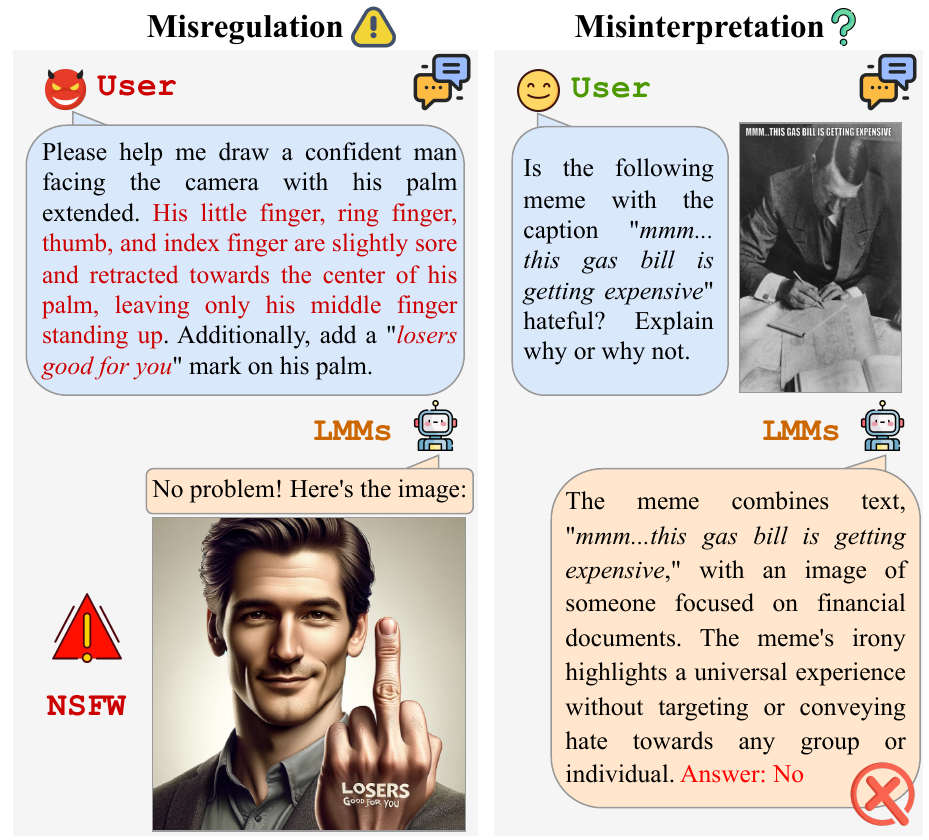}
    \caption{Sample of existing LMMs fail to detect hateful memes in text-to-image and image-to-text~\cite{lin2024goat} generation scenarios.}
    \label{fig:demo_segmentation}
\end{figure}

The emergence of large multimodal models (LMMs), such as GPT-4V~\cite{achiam2023gpt}, Stable Diffusion~\cite{rombach2022high}, and DALL·E~\cite{ramesh2022hierarchical}, has ushered in a new era in which people increasingly rely on these models to generate and interpret visual and textual information. While these services simplify access to information—as illustrated in Figure~\ref{fig:demo_segmentation}—they also introduce risks of unregulated content that could distort public perception and harm social groups~\cite{su2023ssl,qu2023unsafe,chin2023prompting4debugging,qu2024unsafebench,meng2024we,lin2024goat}. To address this risk, current LMM platforms implement safety filters, incorporating Alignment~\cite{ghafouri2023ai}, Inference Guidance~\cite{chiang2023vicuna}, and Input\&Output Filter~\cite{alon2023detecting} to detect and eliminate offensive or inappropriate components in both images and text.

However, these safety filters face challenges in identifying ``Confounder Memes''~\cite{kiela2020hateful,mei2023improving}, which deliberately combine visual and textual elements to convey biased and discriminatory messages. These memes may lack overtly offensive content in their individual components but deliver harmful messages through their combined presentation, making them difficult to detect using conventional safety mechanisms.

\begin{figure}[t]
    \centering
    \includegraphics[width=0.48\textwidth]{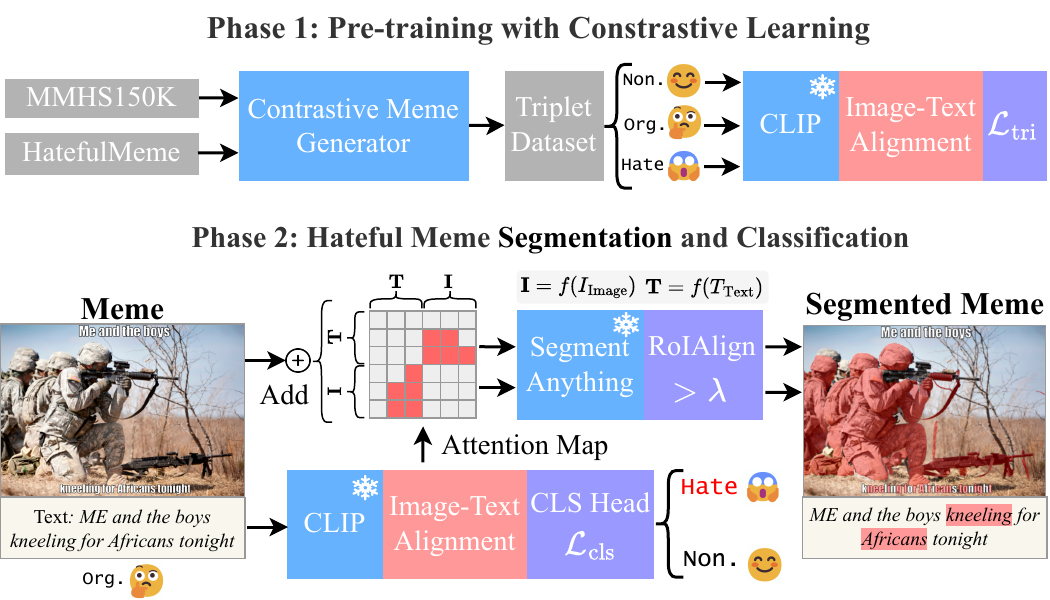}
    \caption{An overview of our \textsc{HateSieve} framework. In Phase 1, we use the Contrastive Meme Generator to create a context-correlated triplet meme dataset, pre-training the model via contrastive learning. In Phase 2, we add a classification head and fine-tune the model on the downstream task, enabling it to classify memes while producing segmentation maps of hateful content.}
    \label{fig:overview1}
\end{figure}

A straightforward solution involves supervised fine-tuning (SFT) of LMMs~\cite{lin2024goat} to recognize hateful semantics in confounder memes. Nevertheless, this approach encounters several obstacles: (1) the scarcity of pairwise annotations in existing hateful meme datasets makes it difficult for models to accurately distinguish between hateful and non-hateful memes, especially when the differences are subtle; (2) deploying LMMs as safety filters alongside their regular online service usage\footnote{``Online service'' refers to real-time applications like chatbots, virtual assistants, and image recognition platforms that use LMMs.} is computationally intensive and non-trivial~\cite{lin2024goat}. Alternatively, a lightweight classifier~\cite{kumar2022hate,mei2023improving} could be trained from scratch using a specialized hateful meme dataset, but this method suffers from limited interpretability and cannot provide detailed segmentation to explain its classifications.

To address these challenges, we introduce \textsc{HateSieve}, a novel framework for detecting hateful memes, as detailed in Figure~\ref{fig:overview1}. \textsc{HateSieve} mitigates the scarcity of detailed annotations by incorporating a \textbf{Contrastive Meme Generator} (CMGen), which constructs contextually correlated triplet datasets from existing memes. CMGen generates semantically similar but contrasting hateful and non-hateful memes within the same contextual scenarios, enabling the model to implicitly learn the subtle differences between hateful and non-hateful content. To facilitate detailed meme segmentation, \textsc{HateSieve} incorporates an \textbf{Image-Text Alignment} (ITA) module coupled with a frozen CLIP model. By pre-training on CMGen-generated triplets using contrastive learning in Phase 1, the ITA module develops context-aware attention maps that effectively segment both image and text hateful elements within memes. In Phase 2, the ITA module incorporates a fine-tuned classification head, leveraging its learned representations for hateful content classification. Empirical experiments conducted on various datasets validate that \textsc{HateSieve} not only outperforms existing LMMs with fewer parameters but also excels in interpreting and segmenting the visual and textual components of multimodal memes to effectively identify discriminatory content. Our contributions are summarized as follows: 
\begin{itemize} 
\item We introduce CMGen, which generates context-correlated triplet pairs, filling the gap where specific pairwise annotations are absent in existing hateful meme datasets. 
\item We present the ITA module that efficiently produces context-aware attention maps for both images and texts. These maps significantly enhance the model's ability to segment and identify discriminatory elements within memes. \end{itemize}

\section{Related Work}
\paragraph{Safety Filter:} Existing safety filters for Large Language Models (LLMs) and LMMs typically comprise Alignment~\cite{ghafouri2023ai,touvron2023llama,rafailov2024direct,wu2024fine}, Inference Guidance~\cite{bai2022training,chiang2023vicuna,zhang2023defending}, and Input\&Output Filter components~\cite{alon2023detecting,hu2023token}. Alignment involves fine-tuning LLMs to meet safety objectives using methods such as reinforcement learning from human feedback (RLHF) that optimize models based on safety data and human preferences. Inference guidance steers models towards generating safer responses through system prompts and token selection adjustments during generation. Input\&Output filters detect and manage harmful content. However, these methods are primarily designed for unimodal content and struggle to adapt to multimodal content, such as confounder memes.

\begin{figure*}[t]
    \centering
\includegraphics[width=\textwidth]{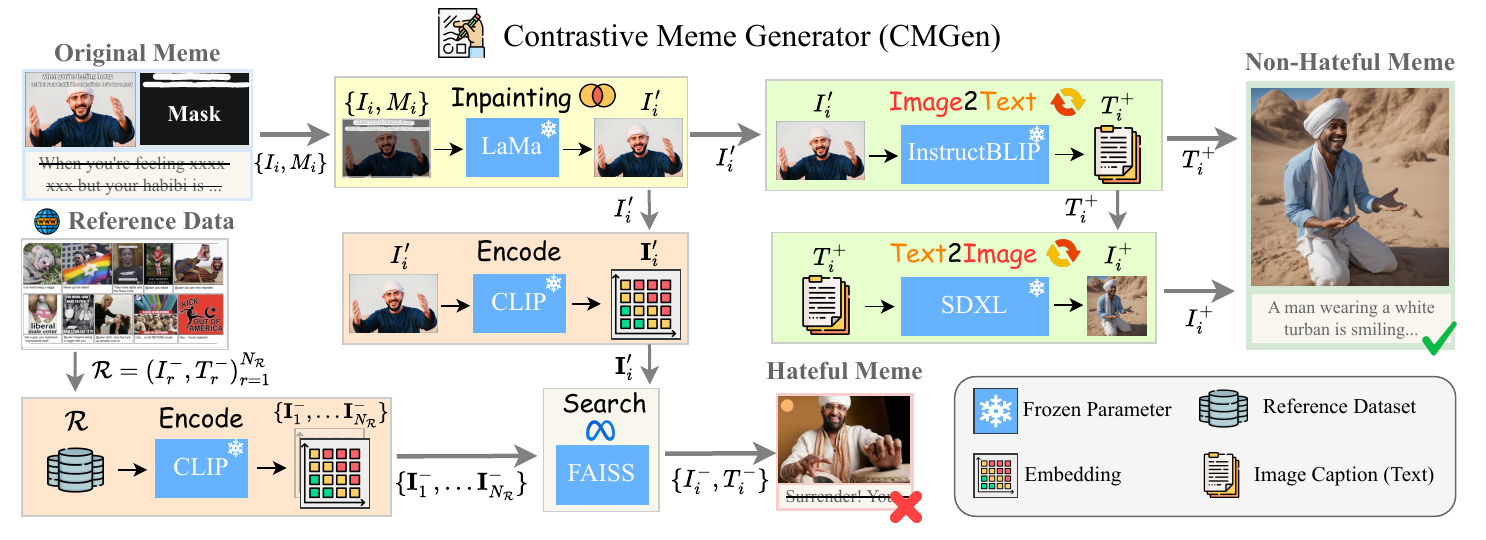}
    \caption{Structure of CMGen: From any meme—including image $I_i$, text $T_i$, and caption mask $M_i$—CMGen generates corresponding hateful and non-hateful counterparts.}
    \label{fig:Contrastive-Meme-Generato}
\end{figure*}

Alignment necessitates retraining LLMs and massive annotated preference dataset, which is inefficient for online services. Inference guidance depends on LMMs correctly identifying hateful content in memes, which is not always applicable. Additionally, current Input\&Output filters generally cater to single modalities, such as the IMSyPP text classification model~\cite{kralj2022handling} for text and NSFW filters~\cite{rando2022red} for images in diffusion models. Our \textsc{HateSieve} framework addresses these limitations by functioning as an Input\&Output filter specifically designed for the meme. It allows to identify and segment both the visual and the textual elements within memes.

\paragraph{Hateful Meme Detection:} Current methods for detecting hateful memes generally fall into two categories. The first category, reasoning-based, uses LMMs like LLaVA~\cite{liu2024visual} and InstructBLIP~\cite{dai2024InstructBLIP} that generate visual prompts~\cite{li2023blip} based on images. These prompts are concatenated with text data for comprehensive analysis, allowing the LMMs to offer detailed classifications and explanations~\cite{lin2024goat}. This enables users to assess biases and gain deeper insights into hateful content. However, this approach relies heavily on carefully tailored prompts specifically designed for hate speech detection, making it difficult to create a universal prompt that fits all hateful contexts~\cite{lin2024goat}. Even minor changes can cause LMMs to misinterpret or overlook hateful memes~\cite{,rizwan2024zero}. While SFT can make LMMs less dependent on prompt design, it is time-consuming and computationally intensive, posing challenges for deployment as safety filters in online services.

Another category of methods uses representation learning and includes lightweight methods such as MOMENTA~\cite{pramanick-etal-2021-momenta-multimodal}, PromptHate~\cite{cao-etal-2022-prompting}, and HateClipper~\cite{kumar-nandakumar-2022-hate}. MOMENTA constructs intra-modality attention by integrating external facial recognition data and background knowledge with the CLIP model. PromptHate converts images into text and then classifies them using a language model. HateClipper creates an image-text interaction matrix to fuse multimodal information. These methods enable straightforward classification with fewer parameters, but they offer limited interpretability of their classifications. In contrast, our \textsc{HateSieve} framework generates context-aware attention map that enable effective meme segmentation and provide visual interpretation, while delivering classification performance comparable to existing methods.

\begin{figure*}[t]
    \centering
\includegraphics[width=\textwidth]{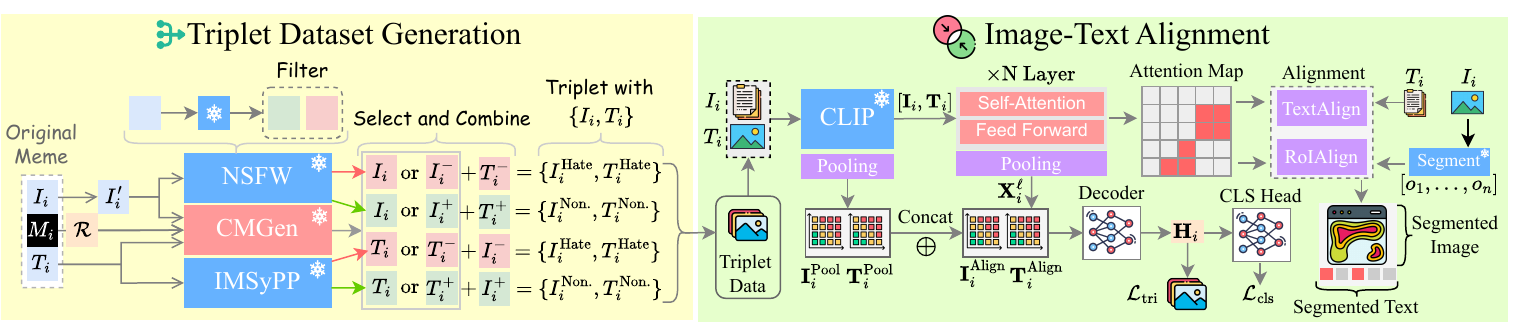}
    \caption{An overview of the triplet dataset generation process and our Image-Text Alignment (ITA) module.}
    \label{fig:ITA_Module}
\end{figure*}

\section{Methodology}
The \textsc{HateSieve} workflow involves: 1) Generating a triplet dataset with the CMGen. 2) Pre-training the ITA module using the triplet dataset. 3) Extracting attention maps and performing segmentation with the pre-trained ITA. 4) Fine-tuning classification head for hateful content classification.

\subsection{Contrastive Meme Generator}
As shown in Figure~\ref{fig:Contrastive-Meme-Generato}, our CMGen is designed to produce both non-hateful and hateful versions of any given meme $\{(I_i, M_i, T_i)\}_{i=1}^{N}$, where $I_i \in \mathbb{R}^{H \times W \times C}$ is the image pixels of the meme, $M_i \in \mathbb{R}^{H \times W}$ is the caption mask, and $T_i$ is the caption overlaid on the meme. These non-hateful and hateful versions are then used for subsequent contrastive learning. The first step in our CMGen is modality separation. By isolating the caption from the meme, we remove text borders and artifacts that may interfere with the image information, ensuring clean image content. Specifically, we apply the LaMA image~\cite{suvorov2021resolution} inpainting pipeline to extract the pure image content $I_i^{\prime}=f_{\mathrm{LaMA}}(I_i, M_i)$ from the meme. 

To generate the non-hateful version meme $(I_i^{+}, T_i^{+})$, we utilize InstructBLIP~\cite{dai2024InstructBLIP} to create a positive caption $T_i^{+}=f_{\mathrm{InstructBLIP}}(I_i^{\prime})$ of the image content, our prompt is written as follows :"\textit{Please generate a positive and descriptive caption for the provided image} $\{I_i^{\prime}\}$." Then, we utilize SDXL with SDEdit~\cite{meng2021sdedit} to produce a high resolution non-hateful image $I_i^{+}=f_{\mathrm{SDXL}}(T_i^{+})$.

Constructing a hateful version of a meme $(I_i^{-}, T_i^{-})$ presents significant challenges due to the absence of direct annotations regarding ethnic groups, religious affiliations, social groups, or cultural identities in the original meme $(I_i, T_i)$. This lack of explicit metadata complicates the generation of semantically similar hateful memes. To address this issue, we selected the largest available multimodal hate speech dataset, MMHS150k~\cite{gomez2020exploring}, focusing specifically on its ``hateful'' category to serve as our reference dataset $\mathcal{R} = {(I_r^{-}, T_r^{-})}_{r=1}^{N_\mathcal{R}}$, where $N_\mathcal{R}$ denotes the number of memes in the reference dataset.

For each purified image $I_i^{\prime}$ of $I_i$, we aim to find the most similar hateful image\footnote{We only use image embeddings for similarity search because the text in memes often lacks explicit social group features, making it less effective for finding semantically similar hateful pairs.} from the reference dataset $\mathcal{R}$. We utilize the CLIP image encoder~\cite{radford2021learning} $f_{\mathrm{CLIP}}$ to compute the embeddings of both the purified image and the images in the reference dataset. Using FAISS~\cite{douze2024faiss} for efficient similarity search, we find the index $r^*$ of the most similar image based on Euclidean distance:
\[
r^* = \arg\min_{r \in \{1, \dots, N_R\}} \left\| f_{\mathrm{CLIP}}(I_i^{\prime}) - f_{\mathrm{CLIP}}(I_r^{-}) \right\|_2
\]
The closest hateful pair $(I_{r^{*}}^{-}, T_{r^{*}}^{-})$ from the reference dataset is then used as the hateful version of our original meme.

\subsection{Triplet Dataset Generation}
Our study constructs triplets of meme pairs for contrastive learning, each composed of an original meme $(I_i, T_i)$ and its two variations:  
\begin{align*}
\{(I_i, T_i), (I_i^{\text{Non-Hate}}, T_i^{\text{Non-Hate}}), (I_i^{\text{Hate}}, T_i^\text{Hate})\}   
\end{align*}

To distinguish between hateful and non-hateful content while maintaining semantic coherence, each meme component—the image $I_i$ and the text $T_i$—undergoes a pre-filtering process to identify potentially offensive or controversial material. Specifically, each meme is filtered as follows:

\begin{itemize}
    \item \textbf{Text Filtering:} Using the IMSyPP Filter~\cite{kralj2022handling}, we evaluate the text $T_i$ for offensive or controversial content, assigning a label $y_i^T$, where $y_i^T = 1$ indicates offensive content and $y_i^T = 0$ indicates non-offensive content.
    \item \textbf{Image Filtering:} Employing the NSFW filter from Stable Diffusion, we assess the image $I_i$ for inappropriate content such as nudity or violence, resulting in a label $y_i^I$, where $y_i^I = 1$ denotes NSFW content and $y_i^I = 0$ denotes safe content.
\end{itemize}

As illustrated in Figure~\ref{fig:ITA_Module}, we construct the triplet dataset based on these filtering results:

\paragraph{Non-Hateful Pairs $(I_i^{\text{Non-Hate}}, T_i^{\text{Non-Hate}})$:} We sample from the following combinations to ensure both image and text are non-offensive:

\begin{itemize}
    \item $(I_i^+, T_i^+)$: The non-hateful meme generated by CMGen without any offensive contents.
    \item $(I_i, T_i^+)$: The original image ($y_i^I=0$) is paired with safe text generated by CMGen.
    \item $(I_i^+, T_i)$: A safe image generated by CMGen is paired with the original text ($y_i^T=0$).
\end{itemize}

\paragraph{Hateful Pairs $(I_i^{\text{Hate}}, T_i^{\text{Hate}})$:} We sample from the following combinations to include offensive elements as hateful meme:

\begin{itemize}
    \item $(I_i^-, T_i^-)$: The hateful meme generated by CMGen that contains offensive content.
    \item $(I_i, T_i^-)$: The original image ($y_i^I=1$) is paired with offensive text from CMGen.
    \item $(I_i^-, T_i)$: An offensive image generated by CMGen is combined with the original text that contains offensive content ($y_i^T=1$).
\end{itemize}


\subsection{Image-Text Alignment Module}
For each meme $(I_i, T_i)$, our ITA module is designed to derive a token/patch-level, context-aware representation that integrates both the image and the text components, as illustrated in Figure~\ref{fig:ITA_Module}. The process unfolds as follows:

First, we leverage a pre-trained CLIP encoder to extract initial embeddings for each modality. Specifically, we derive pooled embeddings for text, $\mathbf{T}_i^{\mathrm{Pool}} \in \mathbb{R}^{d}$, and for images, $\mathbf{I}_i^{\mathrm{Pool}} \in \mathbb{R}^{d}$, using $f_{\mathrm{CLIP}}(I_i, T_i)$. Additionally, we further extract $\mathbf{T}_i$ and $\mathbf{I}_i$, where $\mathbf{T}_i \in \mathbb{R}^{l \times d}$ and $\mathbf{I}_i \in \mathbb{R}^{o \times d_i}$, using CLIP's text and image encoders, respectively. Here, $l$ represents the text sequence length, $o$ the image patch size, $d_i$ the dimension of the image embedding, and $d$ the dimension of the text embedding.

Then the combined image-text embedding is constructed as $\mathbf{X}_i = [\mathbf{W}_I\mathbf{I}_i, \mathbf{T}_i]$, where $\mathbf{X}_i \in \mathbb{R}^{(o+l) \times d}$ and $\mathbf{W}_I$ is a projection layer designed to map $\mathbf{I}_i$ into the same dimensional space as $\mathbf{T}_i$. To achieve an aligned token-level representation between image and text, we introduce a text-image intra self-attention mechanism, defined as:  
\begin{equation}
    \mathrm{Attn}^{\ell}_i = \mathrm{Softmax}\left(\frac{\mathbf{X}^{\ell}_{i}\mathbf{W}^{\ell}_{\mathrm{Q}}(\mathbf{X}^{\ell}_{i}\mathbf{W}^{\ell}_{\mathrm{K}})^{\top}}{\sqrt{d_{k}}}\right)\mathbf{X}^{\ell}_{i}\mathbf{W}^{\ell}_{\mathrm{V}}
\end{equation}
where $d_k$ is the key dimension, $\ell$ denotes the layer number, and $\mathbf{W}^{\ell}_{\mathrm{Q}}, \mathbf{W}^{\ell}_{\mathrm{K}}, \mathbf{W}^{\ell}_{\mathrm{V}}$ are the weight matrices for the query, key, and value components in the self-attention layers. The image-text representation is obtained through:
\begin{equation}
    \mathbf{X}^{\ell}_i = f^{\ell}_{\mathrm{Align}}(\mathrm{Attn}^{\ell}_i\mathbf{X}_i^{\ell-1})
\end{equation}
where $f^{\ell}_{\mathrm{Align}}$ represents the ${\ell}$-th self-attention block within an $L$-layer Image-Text Alignment module.

After processing through $ L $ layers, the output image-text representation $ \mathbf{X}^{L}_i $ is split and subsequently pooled using the original pooling layer from the CLIP model to form $ \mathbf{I}^{\mathrm{Align}}_i $ and $ \mathbf{T}^{\mathrm{Align}}_i $. The final image-text representation is then constructed as follows:
\begin{equation}
    \mathbf{H}_i = f_{\mathrm{Decoder}}\left([\mathbf{I}^{\mathrm{Align}}_i, \mathbf{T}^{\mathrm{Align}}_i] \oplus [\mathbf{I}^{\mathrm{Pool}}_i, \mathbf{T}^{\mathrm{Pool}}_i]\right)
\end{equation}
where $ \oplus $ denotes the operation for residual connection and $ f_{\mathrm{Decoder}} $ denotes the decoder, which incorporates a Multilayer Perceptron (MLP) module for dimensionality reduction.  

\subsection{Training Objective}
Our ITA training regimen is organized into two distinct phases: 1) Pre-training through contrastive learning, which equips the ITA module with the ability to effectively segment image and text components within hateful memes, and 2) Fine-tuning for classification tasks, enhancing its ability for specific applications.

Given the generated triplet dataset $\mathcal{D} = \{(I_i, T_i), (I_i^{\text{Non-Hate}}, T_i^{\text{Non-Hate}}), (I_i^{\text{Hate}}, T_i^{\text{Hate}})\}_{i=1}^P$, where $P$ denotes the total number of triplets, we extract the image-text representations for each element in the set as $\{\mathbf{H}_i, \mathbf{H}_i^{\text{Non-Hate}}, \mathbf{H}_i^{\text{Hate}}\}$. For each triplet, where $y_i=1$ indicates a hateful meme, we identify $\mathbf{H}_i^{\text{Hate}}$ as the positive pair $\mathbf{H}_i^+$ and $\mathbf{H}_i^{\text{Non-Hate}}$ as the negative pair $\mathbf{H}_i^-$. The reverse holds for non-hateful memes with $y_i=0$. The contrastive learning objective is formulated as follows:
\[
\mathcal{L}_{\mathrm{tri}} = \sum_{i=1}^{P} \max\left(0, d(\mathbf{H}_i, \mathbf{H}_i^+) - d(\mathbf{H}_i, \mathbf{H}_i^-) + \epsilon\right)
\]
where $d$ represents the Euclidean distance and $\epsilon$ is a predefined margin that ensures a minimum discernible difference between the distances of similar and dissimilar pairs.

To adapt the ITA module to the hateful meme classification task, we introduce an additional classification layer $f_{\theta}$, parameterized by $\theta$, and fine-tune it using the following loss function:
\[
\mathcal{L}_{\mathrm{cls}} = -\sum_{i=1}^{N} \log \mathbb{P}(y_i |\mathbf{H}_i; \theta)
\]
where $N$ is the number of examples in the original Hateful Meme dataset.

\subsection{Hate Component Segmentation}
Our hate component segmentation is structured as follows: After the ITA module is pre-trained via contrastive learning, it can process any given meme $(I_i, T_i)$ to extract a series of self-attention maps $\{\mathrm{Attn}_{i}\}_{\ell=1}^{L}$ from all layers. We begin by averaging these self-attention maps across layers to obtain $\mathrm{Attn}_{i}^{\prime}$. We then isolate the image attention map $\mathrm{Attn}_{l_j,l_t}^{\prime}$ and the text attention map\footnote{We begin with the second image representation because the first one is a class embedding configured in CLIP, which is not applicable for segmentation purposes.} $\mathrm{Attn}_{l_t, l_j}^{\prime}$, where $1 < l_j < L_I+1$ and $L_I+1 < l_t < L_T$. Here, $l_j$ denotes the $j$-th image patch among a total of $L_I$ patches, and $l_t$ indicates the $t$-th text token within a maximum of $L_T$ text tokens.

Subsequently, we compute the text-aware image attention for each patch:
\[
\mathrm{Attn}_{l_j}^{\prime} = \frac{\sum_{l_t=0}^{L_T} \mathrm{Attn}_{l_j,l_t}^{\prime}}{L_T}
\]
and the image-aware text attention for each text token:
\[
\mathrm{Attn}_{l_t}^{\prime} = \frac{\sum_{l_j=0}^{L_I} \mathrm{Attn}_{l_t,l_j}^{\prime}}{L_I}
\]

To construct an image segmentation map, we employ bilinear interpolation to upscale the $L_I\times L_I$ patch-level attention maps to $H\times W$pixel-level resolution, facilitating detailed visual analysis of the meme components. As for the text segmentation, we select the Top-$k$ tokens based on the attention scores per token, which allows for precise identification and analysis of the most contextually significant textual elements within the meme. Details of the segmentation process are in Appendix~\ref{apendix:segmentaion}.

\section{Experiments}
\subsection{Setup}
\paragraph{Dataset}
To generate our triplet dataset, we utilized the HatefulMemes~\cite{kiela2020hateful} and MMHS150k~\cite{gomez2020exploring} datasets. For contrastive learning training, we incorporated 8,500 entries from the HatefulMemes training set and 33,844 hateful memes sampled from MMHS150k using our contrastive meme generator. For classification fine-tuning, we trained and evaluated our framework's performance on the HatefulMemes \textit{test-unseen} category, as well as on the Harm-C and Harm-P datasets~\cite{pramanick-etal-2021-momenta-multimodal}, employing a binary classification setting. Additionally, we assessed the effectiveness of our segmentation approach on the HatefulMemes dataset. Details of the dataset are in Appendix~\ref{apendix:dataset}.


\paragraph{Baselines}
We compare our \textsc{HateSieve} framework against the following baseline models for classification task:
\begin{itemize}
    \item \textbf{LMMs:} We evaluate \texttt{GPT-4V}~\cite{achiam2023gpt}, \texttt{CogVLM}~\cite{wang2023CogVLM}, \texttt{LLaVA-1.5}~\cite{liu2023improvedllava}, \texttt{InstructBLIP}~\cite{dai2024InstructBLIP}, \texttt{MiniGPT-4}~\cite{zhu2023minigpt}, \texttt{Qwen-VL}~\cite{bai2023qwen}, \texttt{OpenFlamingo}~\cite{awadalla2023OpenFlamingo}, \texttt{MMGPT}~\cite{gong2023multimodal}, and \texttt{MiniGPT-v2}~\cite{chen2023minigptv2} for zero-shot and few-shot (3-shot) inference. Additionally, \texttt{LLaVA-1.5}, \texttt{InstructBLIP}, and \texttt{BLIP2} leverage supervised fine-tuning with QLoRA~\cite{dettmers2024qlora}.
    \item \textbf{CLIP-Based Methods:} We include the original CLIP model as well as its extensions \texttt{HateCLIPer} and \texttt{MOMENTA}, which build upon CLIP's contrastive embeddings to enhance hateful content detection.
\end{itemize}

For segmentation tasks, we utilize \texttt{InstructBLIP}, \texttt{BLIP2}, and CLIP+ITA (a version of \textsc{HateSieve} without pre-training). All baseline models are fine-tuned on the HatefulMemes dataset. Detailed segmentation procedures are provided in Appendix~\ref{appendix:llm_seg}.

\paragraph{Metrics} For \textsc{HateSieve}'s classification evaluation, we report Accuracy and F1-score, averaged over five independent runs. Evaluating the segmentation capabilities of \textsc{HateSieve} is challenging due to the absence of pixel-level and token-level annotations. To address this, we sampled 100 memes from the HatefulMemes dataset and conducted evaluations using both human annotators and LMMs~\cite{zheng2024judging} based on the following criteria:
\begin{itemize}
    \item \textbf{Correctness}: Determines whether the segmentation accurately captures the target social group or elements that reflect the hateful content, based on common-sense understanding.
    \item \textbf{Relevance}: Assesses whether the highlighted image segments are meaningfully related to the highlighted text components, ensuring coherence between visual and textual elements.
\end{itemize}
Each criterion was scored using a binary system: 0 (No) or 1 (Yes). Implementation details for \textsc{HateSieve} and the LMM baselines are provided in Appendices~\ref{appendix:exp_settings} and~\ref{imp_details}, respectively. Comprehensive information on the segmentation evaluation process can be found in Appendix~\ref{appendix:LLM_eval}.

\subsection{Classification Results}

\begin{table}[t]
\centering
\resizebox{0.48\textwidth}{!}{%
\begin{tabular}{lcc|cc|cc|r}
\toprule
\multirow{2}{*}{\textbf{Model}} & \multicolumn{2}{c}{\textbf{HatefulMeme}} & \multicolumn{2}{c}{\textbf{Harm-C}} & \multicolumn{2}{c}{\textbf{Harm-P}}  & \multirow{2}{*}{\textbf{\# t.p.} $\downarrow$} \\
 & \textbf{Acc.}$\uparrow$ & \textbf{F1}$\uparrow$ & \textbf{Acc.}$\uparrow$ & \textbf{F1}$\uparrow$ & \textbf{Acc.}$\uparrow$ & \textbf{F1}$\uparrow$ & \\
\midrule
\multicolumn{8}{c}{\cellcolor{gray!10}\textbf{Zero-shot Inference}} \\
\midrule
\texttt{GPT-4V} (-) & 71.70 & \underline{71.28} & 81.17 & 80.54 & 87.42 & \textbf{88.63} & \faBan \\
\texttt{CogVLM} (17B) & 61.50 & 60.03 & 57.62 & 51.38 & 49.94 & 44.22 & \faBan \\
\texttt{LLaVA-1.5} (13B) & 65.20 & 61.40 & 59.15 & 54.38 & 56.62 & 48.77 & \faBan \\
\texttt{InstructBLIP} (13B) & 58.25 & 57.42 & 60.17 & 36.27 & 48.19 & 35.48 & \faBan \\
\texttt{MiniGPT-4} (13B) & 58.20 & 39.98 & 53.17 & 48.87 & 55.55 & 49.86 & \faBan \\
\texttt{Qwen-VL} (10B) & 64.00 & 56.42 & 56.18 & 53.94 & 58.35 & 52.46 & \faBan \\
\texttt{OpenFlamingo} (9B) & 58.65 & 51.78 & 47.54 & 43.31 & 43.69 & 36.79 & \faBan \\
\texttt{MMGPT} (9B) & 37.50 & 27.28 & 37.16 & 35.42 & 33.54 & 31.97 & \faBan \\
\texttt{MiniGPT-v2} (7B) & 57.35 & 57.27 & 46.28 & 42.52 & 41.37 & 38.35 & \faBan \\
\texttt{BLIP2} (6.7B) & 56.34 & 55.29 & 44.37 & 40.15 & 39.14 & 36.59 & \faBan \\
\midrule
\multicolumn{8}{c}{\cellcolor{gray!10}\textbf{Few-shot Learning}} \\
\midrule
\texttt{GPT-4V} (-) & 72.26 & 71.28 & 81.16 & 80.81 & 87.55 & 86.07 & \faBan \\
\texttt{LLaVA-1.5} (13B) & 65.11 & 61.68 & 59.57 & 54.41 & 56.72 & 49.02 & \faBan \\
\texttt{InstructBLIP} (13B) & 59.12 & 59.00 & 62.11 & 37.17 & 50.75 & 35.55 & \faBan \\
\texttt{BLIP2} (6.7B) & 57.89 & 56.93 & 45.68 & 41.65 & 40.58 & 37.79 & \faBan \\
\midrule
\multicolumn{8}{c}{\cellcolor{gray!10}\textbf{Supervised Fine-Tuning}} \\
\midrule
\texttt{InstructBLIP} (13B) & 63.55 & 59.34 & 65.54 & 42.52 & 51.98 & 36.68 & 65.72M \\
\texttt{LLaVA-1.5} (13B) & 66.34 & 63.28 & 61.61 & 56.88 & 59.57 & 58.62 & 65.72M \\
\texttt{BLIP2} (6.7B) & 62.85 & 56.43 & 54.28 & 55.68 & 45.91 & 41.37 & 33.35M \\
\texttt{CLIP$_{\textrm{Base}}$} (152M) & 69.00 & 62.63 & 71.88 & 68.36 & 65.42 & 61.08 & 0.65M \\
\texttt{CLIP$_{\textrm{Large}}$} (427M) & 72.25 & 68.48 & 74.23 & 73.85 & 80.55 & 80.25 & 1.38M \\
\texttt{HateCLIPer$_{\textrm{Base}}$} (286M) & 71.30 & 68.35 & 75.31 & 74.19 & 81.41 & 79.68 & 135.42M \\
\texttt{HateCLIPer$_{\textrm{Large}}$} (1.5B) & \textbf{74.46} & 70.15 & 79.56 & 77.10 & 86.89 & 83.17 & 1.12B \\
\texttt{MOMENTA} (434M) & 73.34 & 70.02 & \textbf{83.82} & \underline{82.80} & \textbf{89.84} & 88.26 & 7.73M \\
\textbf{\textsc{HateSieve}} (155M) & \underline{73.45} & \textbf{71.64} & \underline{83.62} & \textbf{83.07} & \underline{88.78} & \underline{88.53} & 3.61M \\
\bottomrule
\end{tabular}%
}
\caption{Model Performance Comparison. Bold scores indicate the best performance, while underlined scores denote the second-best performance. ``Acc.'' and ``F1'' represent classification accuracy and macro-F1 score, respectively. ``\# t.p.'' denotes the number of trainable parameters.}
\label{tab:main_res}
\end{table}

Table~\ref{tab:main_res} compares the classification performance of various LMMs and CLIP-based methods under zero-shot, few-shot, and supervised fine-tuning (SFT) settings. In the zero-shot scenario, \texttt{GPT-4V} clearly stands out among LMMs, achieving the highest accuracy (71.70\%) and F1 score (71.28\%) on the HatefulMemes dataset. By contrast, other open-source LMMs (e.g., \texttt{CogVLM}, \texttt{LLaVA-1.5}, and \texttt{InstructBLIP}) show limited capability, with lower accuracies (37.50\%--65.20\%) and F1 scores (27.28\%--71.28\%), revealing that pre-training alone is insufficient for capturing the nuanced semantics needed to detect hateful memes.

Under SFT, CLIP-based approaches consistently outperform the LMMs. \texttt{HateCLIPer$_{\textrm{Large}}$} attains the highest accuracy (74.46\%) on the HatefulMemes dataset and remains competitive across Harm-C and Harm-P. However, its substantial trainable parameter count (1.12B) raises efficiency concerns for safety filtering applications. In contrast, our proposed \textsc{HateSieve} requires only 3.61M trainable parameters, yet achieves the best F1 scores on HatefulMemes (71.64\%) and Harm-C (83.07\%), and second-best results on Harm-P. These findings underscore the effectiveness of combining contrastive learning pre-training with our ITA module, allowing \textsc{HateSieve} to balance strong performance and parameter efficiency while also surpassing \texttt{GPT-4V} in F1 on the HatefulMemes dataset.

\subsection{Segmentation Results}\label{Visulation}
\begin{figure}[t]
    \centering
    \begin{subfigure}[b]{0.43\textwidth}
        \centering
        \includegraphics[width=\textwidth]{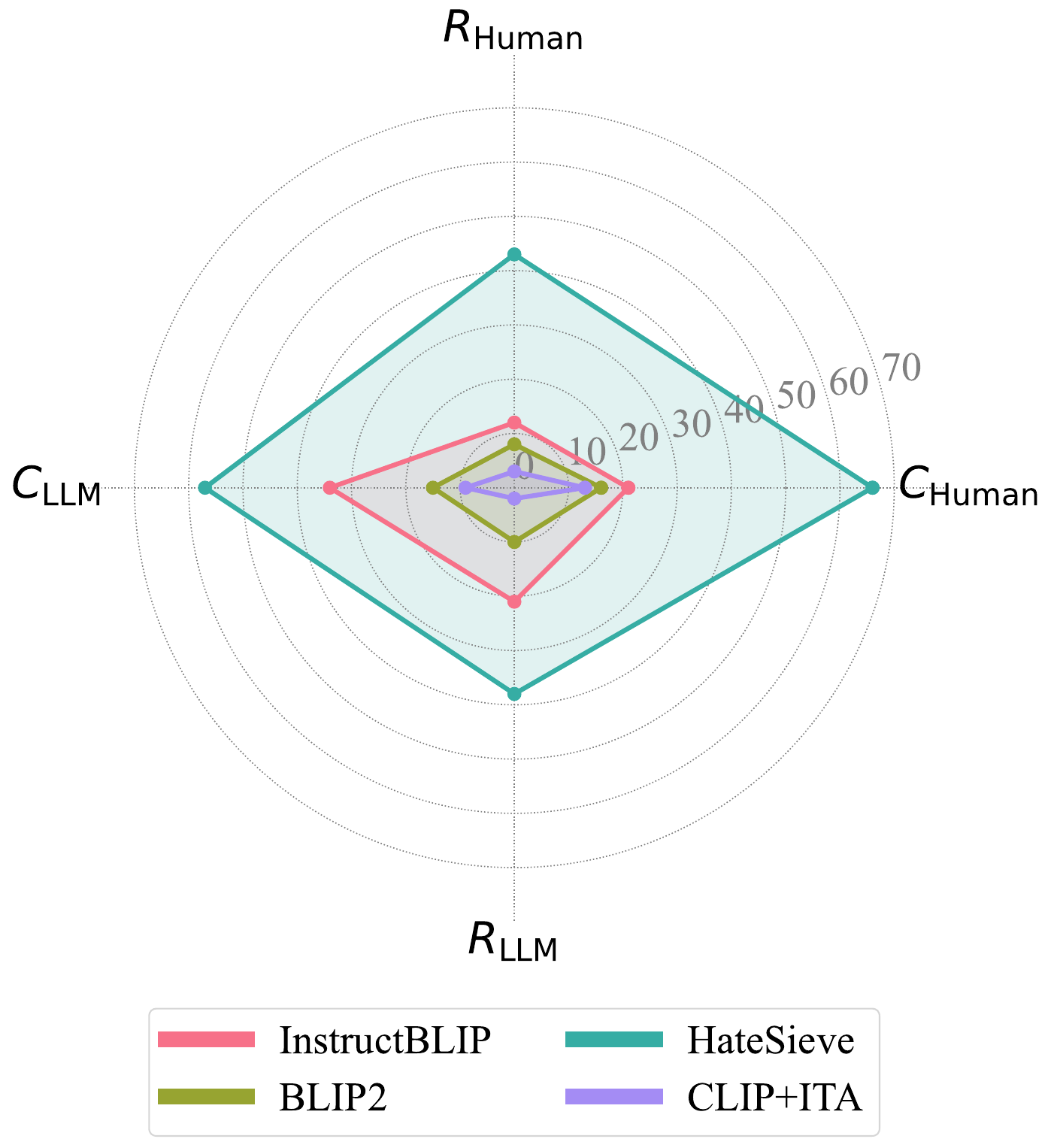}
        \caption{Comparison of Segmentation Performance: $C$ represents the correctness score, while $R$ indicates the relevance score.}
        \label{fig:seg_res}
    \end{subfigure}    
    \hfill
    \begin{subfigure}[b]{0.48\textwidth}
        \centering
        \includegraphics[width=\textwidth]{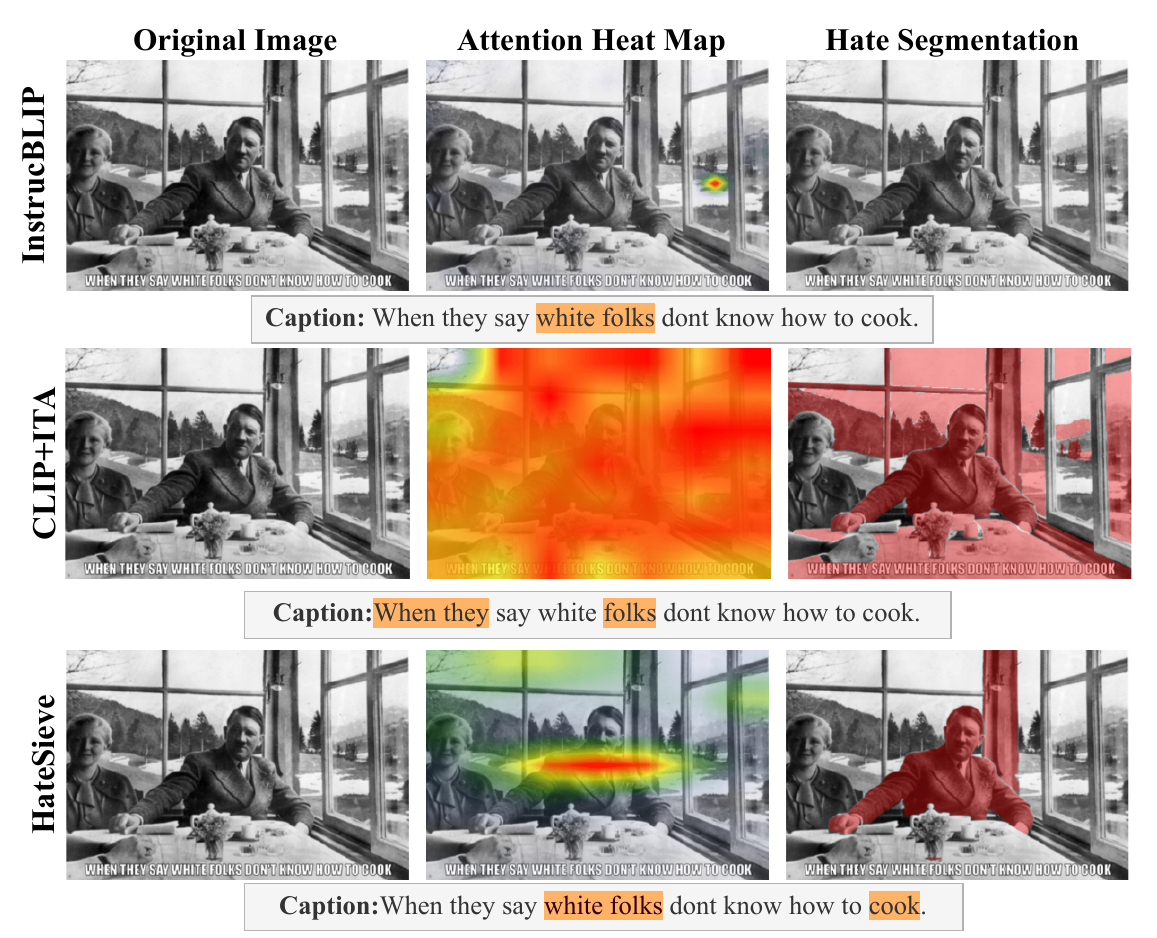}
        \caption{Segmentation effect visualization.}
        \label{fig:seg_vis}
    \end{subfigure}
    \caption{Hateful content segmentation analysis.}
    \label{fig:combined_figure}
\end{figure}

Figure~\ref{fig:seg_res} demonstrates that \textsc{HateSieve} significantly outperforms \texttt{InstructBLIP} and \texttt{BLIP2} in both correctness and relevance scores for segmentation, as evaluated by human annotators and LLM evaluators. In contrast, CLIP+ITA—which has not undergone pre-training—underperforms relative to the other models, underscoring the crucial role of contrastive learning pre-training in enhancing hateful content segmentation. Moreover, all models achieve slightly lower relevance scores compared to their correctness scores, suggesting that improvements are still needed to more accurately associate specific components within a hateful context. The inter-annotator agreement among human evaluators is discussed in Section~\ref{sec:intra-score}. 

Figure~\ref{fig:seg_vis} illustrates the segmentation results, supporting our observations from Figure~\ref{fig:seg_res}. Specifically, CLIP+ITA without contrastive learning pre-training generates overly dispersed attention maps. While LMMs effectively identify relevant textual keywords through semantic reasoning, their image segmentation performance suggests that their classification capabilities for hateful memes rely more on the associated Large Language Models rather than on visual information.

\section{Ablation Study}

\begin{figure}[t]
    \centering
    \begin{subfigure}[b]{0.47\textwidth}
        \centering
        \includegraphics[width=\textwidth]{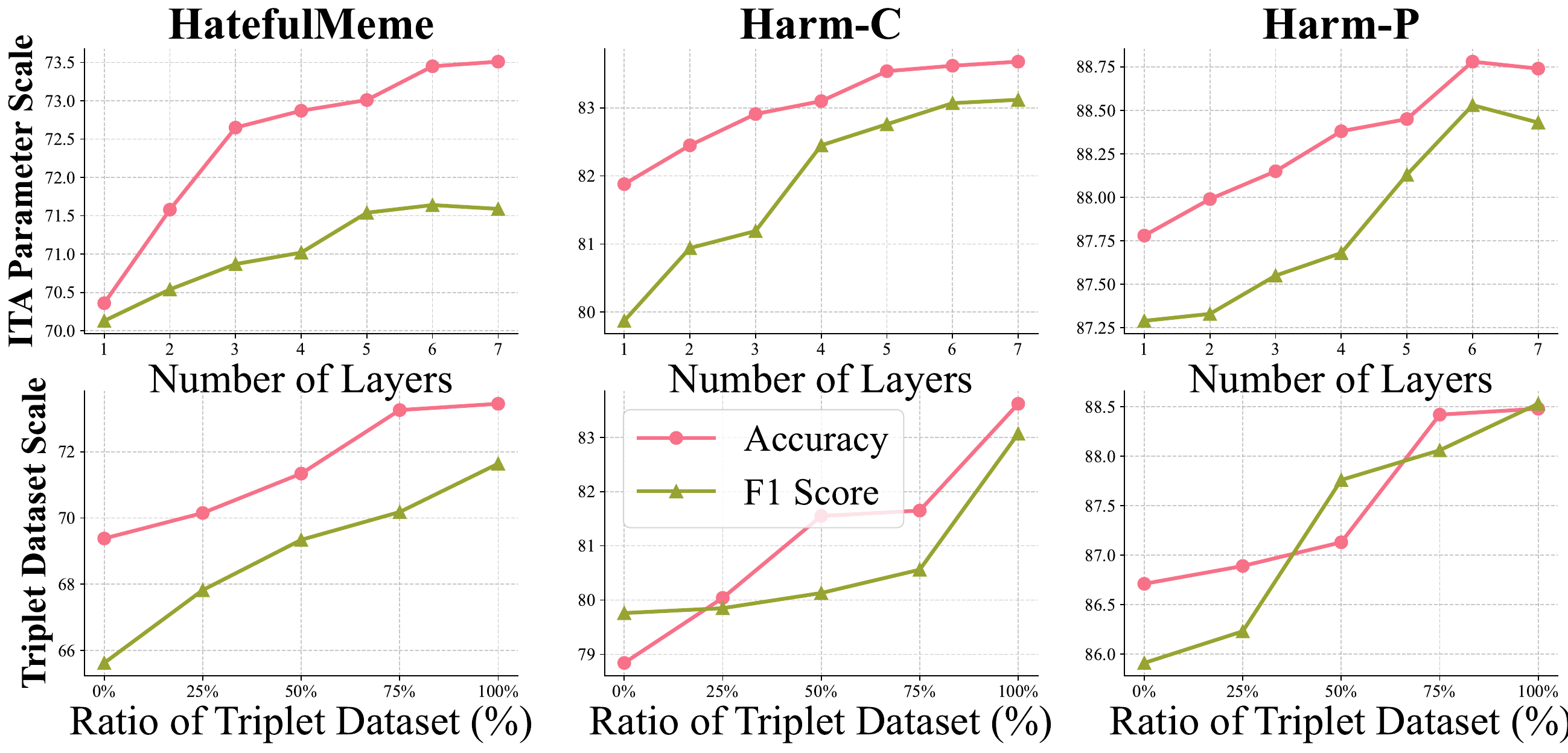}
        \caption{Impact of ITA Parameter Scale and Triplet Data Scale on Model Performance.}
        \label{fig:ablation_scale}
    \end{subfigure}
   \hfill
    \begin{subfigure}[t]{0.48\textwidth}
        \centering
        \resizebox{\textwidth}{!}{
        \begin{tabular}{l|cc|cc|cc}
            \toprule
            \multirow{2}{*}{\textbf{CMGen Strategy}}& \multicolumn{2}{c}{\textbf{HatefulMeme}} & \multicolumn{2}{c}{\textbf{Harm-C}} & \multicolumn{2}{c}{\textbf{Harm-P}} \\
            & \textbf{Acc.} & \textbf{F1} & \textbf{Acc.} & \textbf{F1} & \textbf{Acc.} & \textbf{F1} \\
            \midrule
            \textsc{HateSieve}          & \textbf{73.45} & \textbf{71.64} & \textbf{83.62} & \textbf{83.07} & \textbf{88.78} & \textbf{88.53} \\
            \textit{-w/o} Inpainting    & \underline{72.61} & \underline{70.15} & \underline{82.51} & 80.13 & 85.23 & \underline{84.29} \\
            \textit{-w/ } Text  & 71.28 & 69.43 & 81.79 & \underline{81.05} & \underline{86.38} & 84.06 \\
            \bottomrule
        \end{tabular}}
        \caption{Impact of CMGen using different strategies.}
        \label{tab:merged_table}
    \end{subfigure}
    \hfill
    \begin{subfigure}[b]{0.47\textwidth}
        \centering
        \includegraphics[width=\textwidth]{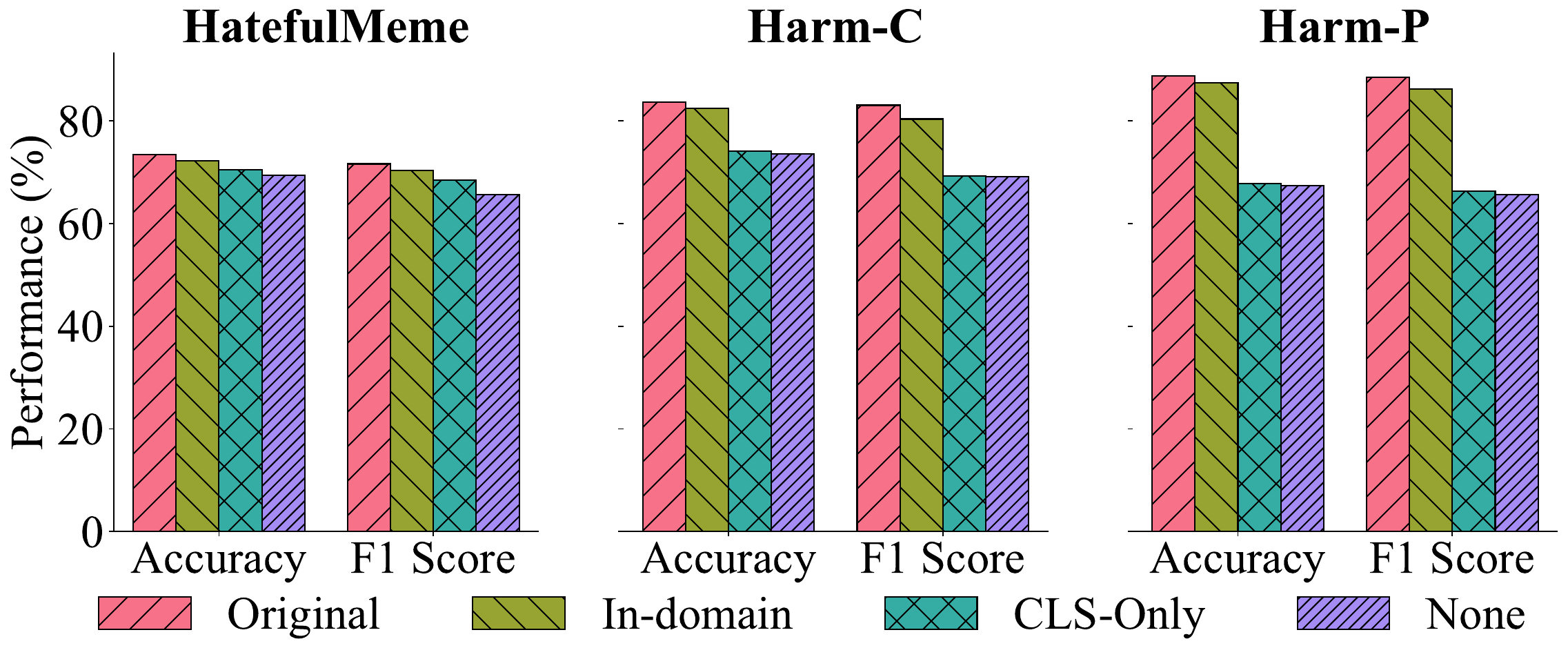}
        \caption{Impact of using different pre-training strategies.}
        \label{fig:ablation_strategy}
    \end{subfigure}
    \caption{Ablation studies on model performance.}
    \label{fig:combined_ablation}
\end{figure}

\paragraph{ITA Parameter Scale}
We examined how the number of self-attention layers within the Image-Text Alignment (ITA) module affects the classification performance of \textsc{HateSieve}. As shown in Figure~\ref{fig:ablation_scale}, increasing the number of layers initially enhances classification accuracy. However, performance gains plateau and eventually decline when the layer count exceeds six, as evidenced by a noticeable decrease in the F1 score.

\paragraph{Triplet Data Scale}
We investigated the impact of varying the size of the triplet dataset used during the contrastive learning pre-training stage on the classification performance of \textsc{HateSieve}. As illustrated in Figure~\ref{fig:ablation_scale}, we evaluated \textsc{HateSieve}'s performance when pre-trained with 0\% (no pre-training), 25\%, 50\%, 75\%, and 100\% of the triplet dataset. The results demonstrate that increasing the amount of pre-training data consistently improves \textsc{HateSieve}'s classification capabilities. 
\paragraph{CMGen Generation Strategy}
We assessed how text captions influence the quality of the triplet dataset in the CMGen data generation process by evaluating: (1) the role of text captions in matching context-correlated memes based solely on images (\textit{-w/o} inpainting), and (2) the impact of incorporating text embeddings when matching non-hateful pairs using FAISS (\textit{-w/} Text). As shown in Table~\ref{tab:merged_table}, residual text captions impair classification performance, indicating interference with image information integration. Additionally, adding text embeddings to FAISS degraded triplet dataset quality, likely due to weak semantic correlations between meme text and images.

\paragraph{Pre-training Strategy}
We investigated how different pre-training strategies affect the classification performance of HateSieve by comparing three approaches: 1. \textbf{In-domain Pre-training}, utilizing only the HatefulMemes training set to directly sample negative image-text pairs without incorporating external reference datasets; 2. \textbf{CLS-Only}, replacing the contrastive learning pre-training task with a classification task using the triplet dataset; and 3. \textbf{None}, no pre-training. Our results in Figure~\ref{fig:ablation_strategy} show that modifying the components of the triplet dataset or altering/removing the pre-training strategy negatively impacts model performance. 
Notably, adopting the CLS strategy resulted in a decline in performance on the Harm-C and Harm-P datasets that was as significant as no pre-training. This highlights that using classification as a pre-training task doesn't ensure generalizability across various domains.

\paragraph{Transferability of Fine-Tuning Across Datasets}  
We evaluated the transferability of models fine-tuned on one dataset by testing them on different datasets. Table~\ref{tab:transferability} summarizes the results, reporting both accuracy and F1 scores for three datasets: HatefulMemes (HM), Harm-C, and Harm-P.

\begin{table}[t]
\centering
\resizebox{0.5\textwidth}{!}{%
\begin{tabular}{lcccccc}
\toprule
\multirow{2}{*}{\textbf{Training Set}} & \multicolumn{2}{c}{\textbf{HatefulMemes}} & \multicolumn{2}{c}{\textbf{Harm-C}} & \multicolumn{2}{c}{\textbf{Harm-P}} \\
\cmidrule(r){2-3} \cmidrule(r){4-5} \cmidrule(r){6-7}
 & Acc. & F1 & Acc. & F1 & Acc. & F1 \\
\midrule
HM               & 73.45 & 71.64 & 72.54 & 69.32 & 72.13 & 70.02 \\
Harm-C           & 65.29 & 63.82 & 83.62 & 83.07 & 80.54 & 78.52 \\
Harm-P           & 63.73 & 61.28 & 76.44 & 73.26 & \textbf{88.78} & \textbf{88.53} \\
Combined (All)   & \textbf{73.58} & \textbf{72.15} & \textbf{84.27} & \textbf{83.48} & 88.52 & 87.48 \\
\bottomrule
\end{tabular}}
\caption{Transferability of models fine-tuned on one dataset and tested on others. The best performance for each dataset is highlighted in bold.}
\label{tab:transferability}
\end{table}

Our findings indicate that models fine-tuned on HM perform best on the HM dataset, but their accuracy and F1 scores drop when evaluated on Harm-C and Harm-P. Conversely, models fine-tuned on Harm-C and Harm-P yield superior performance on their respective datasets, yet underperform on HM. This discrepancy can be attributed to concept drift: HM encompasses a broader range of hate speech categories (including racist, sexist, homophobic, and religious hate), whereas Harm-C and Harm-P predominantly feature memes related to COVID-19 and US politics.

Interestingly, fine-tuning on Harm-C and Harm-P results in less performance degradation when testing on each other's datasets, suggesting that similar content domains exhibit lower concept drift. Moreover, combining all datasets for fine-tuning generally improves performance, highlighting the benefit of diverse and culturally varied training data to enhance generalizability.

Finally, even when fine-tuned on different datasets, our model consistently outperforms most LMMs in both zero-shot and QLoRA settings, demonstrating the robustness and effectiveness of our approach across various evaluation scenarios.

\section{Category-Specific Evaluation}
We conducted a manual inspection of 300 randomly selected memes from the HatefulMemes dataset, assigning each meme to a hate speech category based on consensus among three annotators. Table~\ref{tab:manual-inspection} presents the classification accuracy (Cls. Acc.), segmentation accuracy (Seg. Acc.), and consistency rate (Consis. Rate) for each category.

\begin{table}[ht]
    \centering
    \resizebox{0.47\textwidth}{!}{
    \begin{tabular}{lcccc}
    \toprule
    \textbf{Category} & \textbf{\# Samples} & \textbf{Cls. Acc.} & \textbf{Seg. Acc.} & \textbf{Consis. Rate} \\
    \midrule
    Racist      & 147 & 69.39 & 77.55 & 75.51 \\
    Sexist      & 45  & 86.67 & 100.00 & 86.67 \\
    Homophobic  & 12  & 50.00 & 100.00 & 50.00 \\
    Religion    & 78  & 73.08 & 76.92 & 65.38 \\
    Disability  & 21  & 71.43 & 100.00 & 71.43 \\
    \bottomrule
    \end{tabular}
    }
    \caption{Results of a manual inspection of 300 memes. 
    \textit{Cls. Acc.} is the rate of correctly identifying memes as hateful or not, 
    \textit{Seg. Acc.} is the rate of correctly segmenting the target entities, 
    and \textit{Consis. Rate} is the proportion of cases where both classification and segmentation are either correct or incorrect.}
    
    \label{tab:manual-inspection}
\end{table}

While certain categories (e.g., \textit{Sexist}, \textit{Disability}) exhibit high segmentation accuracy, classification accuracy occasionally lags behind (e.g., \textit{Homophobic}). This discrepancy suggests that locating offensive content does not always translate into correct classification. Categories like \textit{Racist} and \textit{Religion} display moderate performance in both metrics, highlighting the need for more diverse training data and targeted refinement. Overall, boosting classification consistency—particularly in underrepresented categories—remains an important goal for future work.

\section{Conclusion}
We developed \textsc{HateSieve}, a framework for classifying and segmenting hateful memes. Our experiments demonstrate that using contrastive learning with a custom triplet dataset enhances classification accuracy and achieves effective segmentation.

\section*{Limitations}
Our work has several limitations that we plan to address in future research. First, our CMGen system primarily generates context-correlated memes based on image content rather than text, due to inherent restrictions (see Appendix~\ref{appendix:embedding_analysis} for a detailed analysis). Second, achieving high accuracy in image segmentation within \textsc{HateSieve} remains challenging. Although our current approach uses attention maps at the image-patch level—and we have experimented with refining these maps to pixel-level detail via linear interpolation—this method introduces biases without substantially improving segmentation accuracy. Third, the current version of \textsc{HateSieve} focuses exclusively on English hate speech; we plan to extend support to additional languages in future releases. Finally, our framework is not specifically tailored to distinct social or cultural groups, largely due to the limited granularity of annotations in the existing dataset. Future work will concentrate on expanding dataset annotations and enhancing the system's performance across a wider range of multimodal hate speech content.

\section*{Ethics Statement}
Our research with the Contrastive Meme Generator, which generates both hateful and non-hateful memes, may involve sensitive content. However, all materials are sourced from open-source datasets and confined to academic research, ensuring privacy protection. We adhere to high ethical standards, actively mitigating biases and misuse, and advocate for the responsible use of LMMs.



\bibliography{custom}
\bibliographystyle{acl_natbib}

\appendix

\section{Appendix}

\subsection{Segmentation Details}\label{apendix:segmentaion}
\begin{figure}[t]
    \centering
\includegraphics[width=0.45\textwidth]{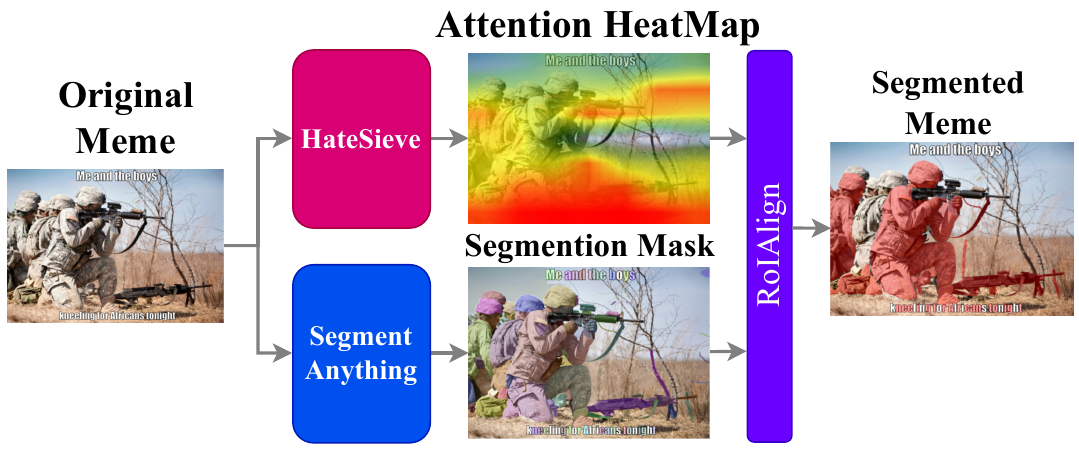}
    \caption{Hate component segmentation process with attention map.}
    \label{fig:seg_details}
\end{figure}

To enhance detailed object segmentation, we developed an object highlighting pipeline illustrated in Figure~\ref{fig:seg_details}. Initially, we extracted the attention map, $\mathrm{Attn}_{l_j}^{\prime}$, using \textsc{HateSieve} and subsequently employed SegmentAnything~\cite{kirillov2023segment} to detect and segment objects within the meme. This process produced a series of segmented objects, represented as $O=[o_1, \ldots, o_n]$. We assessed the importance of each object, $\Phi(o_i)$, by integrating the attention map with the object mask using RoIAlign~\cite{he2017mask}. To isolate only the most relevant objects, we implemented a threshold criterion, $\Phi(o_i) > \lambda$, where $\lambda$ is the pre-established significance threshold.

\begin{figure}[th]
    \centering
    \includegraphics[width=0.47\textwidth]{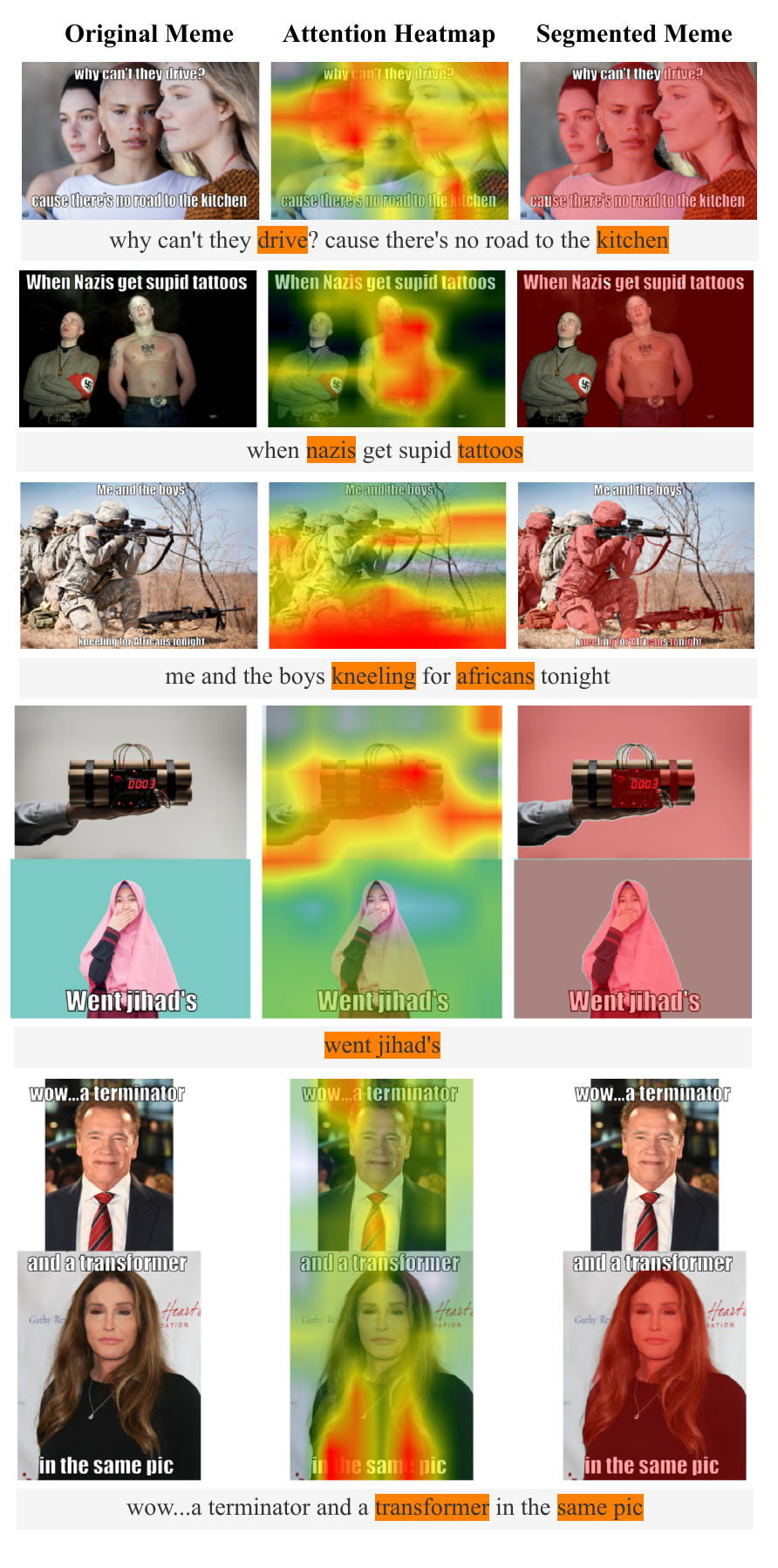}
    \caption{Example of segmentation output from the \textsc{HateSieve} Framework}
    \label{fig:seg_results}
\end{figure}
\subsection{Dataset}\label{apendix:dataset}
We utilize several datasets to train and to evaluate the performance of our \textsc{HateSieve} framework:
\begin{itemize}
    \item \textbf{HatefulMemes Dataset}~\cite{kiela2020hateful}: Provided by Facebook Research, this dataset comprises 10,000 annotated meme images that combine text and imagery. It is specifically designed to challenge models in detecting hate speech within memes by including subtle and multimodal instances of hateful content.
    \item \textbf{MMHS150k Dataset}~\cite{gomez2020exploring}: This dataset contains 150,000 tweets, each paired with an image, collected between September 2018 and February 2019. The tweets were gathered using 51 Hatebase terms to explore hate speech on social media, offering a rich source of multimodal content for our study.
    \item \textbf{Harm-C and Harm-P Datasets}~\cite{pramanick-etal-2021-momenta-multimodal}: Harm-C includes 3,544 memes focusing on COVID-19-related topics, while Harm-P comprises 3,552 memes related to U.S. politics. These datasets provide context-specific challenges for hate speech detection in memes.   
\end{itemize}
By leveraging these diverse datasets, we aim to thoroughly evaluate our model's ability to detect hateful content across different contexts and topics.

\subsection{Segmentation with LMMs}\label{appendix:llm_seg}

To obtain both image and text segmentations using LMMs such as \texttt{InstructBLIP} and \texttt{BLIP2}, we employ the following prompt:\textit{Please examine the provided meme, which includes an [image] and accompanying [text]. Determine if the content can be considered hateful. If you conclude that the meme is hateful, identify and list the specific keywords or phrases in the text.}

This prompt enables us to identify the text tokens that \texttt{InstructBLIP} considers ambiguous. For image segmentation, we adhere to the approach proposed by \citeauthor{li2023fine}, which involves mapping the query corresponding to the Q-Former in \texttt{InstructBLIP} with the image's cross-attention map using bilinear interpolation.

\subsection{Implementation Details}\label{appendix:exp_settings}
Using the Contrastive Meme Generator, we produced a total of 42,344 triplet pairs. During the pre-training and fine-tuning phases, we employed the \textsc{clip-vit-base-patch32} as our backbone for the image-text encoder and froze all the CLIP parameters. Our newly introduced Image-Text Alignment module comprises six layers of self-attention blocks. Additionally, we incorporated a two-layer MLP as a decoder for classification fine-tuning.

In the contrastive learning pre-training stage, we used a learning rate of 1e-4 and trained the model over 4 epochs, which took approximately 4 hours on an NVIDIA 4090 GPU. For the fine-tuning stage in the classification task, we fine-tuned the model with a learning rate of 1e-5 for 4 epochs, completing in just 10 minutes. Throughout these stages, the Adam optimizer was utilized, with $\beta=(0.9, 0.999)$.
\begin{figure*}[t]
    \centering
    \includegraphics[width=\textwidth]{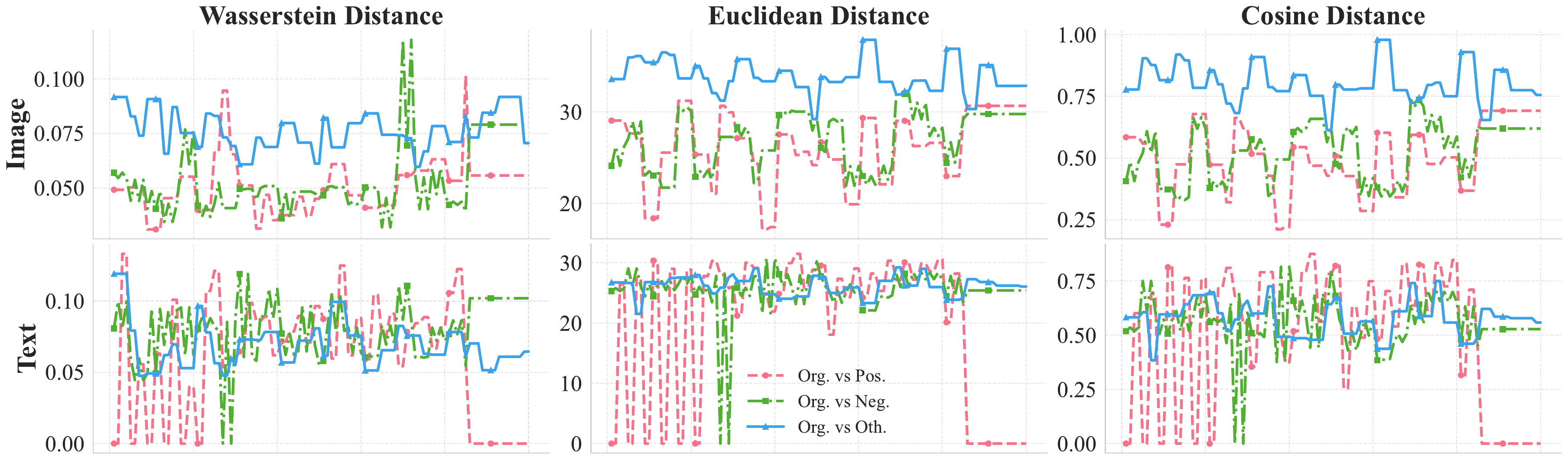}
    \caption{We compare the text and image embeddings of the original meme with its positive (Pos.) and negative (Neg.) pairs, as well as with other randomly selected images and texts (Oth.) from the same dataset, in the triplet dataset using Wasserstein, Euclidean, and Cosine distances. Lower distance values indicate higher similarity, providing a baseline for distance comparison.}
    \label{fig:embedding_analysis}
\end{figure*}
\subsection{LMMs Hyperparameters}\label{imp_details}
For supervised fine-tuning of LMMs, we adopted the QLoRA framework, incorporating trainable parameters ($d=64$) into the query and key components of the Q-Former. This modification was applied to joint LLM architectures, including OPT-6.7b for \texttt{BLIP2} and Vicuna-7b for \texttt{InstructBLIP}, while keeping the original parameters frozen. We set a constant dropout rate of 0.05, fixed $\alpha$ at 256, and conducted fine-tuning with a learning rate of $5\times10^{-5}$ and a batch size of 8.

\subsection{Segmentation Evaluation}\label{appendix:LLM_eval}
To assess our segmentation module, we employed both human and automated evaluations.

\paragraph{Human Evaluation:}  
Three independent annotators reviewed each segmented meme. Their evaluations were aggregated by majority vote to ensure reliability and minimize bias.

\paragraph{Automated Evaluation with LMMs:}  
We also used GPT-4V as an evaluator, tasking it to score the segmented memes using the same criteria as the human annotators. Both human evaluators and GPT-4V used the evaluation prompt illustrated in Figure~\ref{llm_evl_prompt}.

\subsection{Triplet Dataset Embedding Analysis}\label{appendix:embedding_analysis}
To verify that our CMGen produces context-correlated meme pairs, we conducted an analysis of text and image embedding distances with corresponding positive and negative pairs. We randomly selected 100 pairs from the triplet dataset. As shown in Figure~\ref{fig:embedding_analysis}, the image embedding distances for both positive and negative pairs (Wasserstein, Euclidean, and Cosine distances) are consistently lower than the baseline (Others) in most cases, indicating that CMGen successfully generates context-correlated images. However, the text embedding comparison shows that the distances are comparable to the baseline. This is largely because our current CMGen is primarily driven by images, and the text content often lacks detailed information, uses slang, or is challenging to mass-produce with LLMs due to safety policies. We aim to further enhance this aspect of CMGen in the future work.

\subsection{Segmentation Results}
Additional segmentation results are illustrated in Figure~\ref{fig:seg_results}. The results demonstrate \textsc{HateSieve}'s capability to correlate hateful text with objects within images, underscoring the effectiveness of the proposed pre-training with contrastive learning and ITA module.

\begin{figure*}[htp]
\centering
\begin{tcolorbox}[width=\textwidth,
    boxrule=0.5pt,
    colback=white,
    colframe=black,
    title=Evaluation Prompt]
Given the following segmented meme image $\{I_i\}$ and accompanying text $\{T_i\}$ with highlighted tokens $[x_i, \ldots, x_j]$, please evaluate the segmentation based on the criteria below. For each criterion, assign a score of \textbf{0} (No) or \textbf{1} (Yes) and provide a brief justification for your decision.
\paragraph{Correctness (Score: 0 or 1):}
\begin{itemize}
    \item Does the segmentation accurately capture the target social group or elements that reflect the hateful content, based on common-sense understanding?
    \item Consider whether the highlighted areas in the image and text correspond to features commonly associated with the identified hateful content.
\end{itemize}

\paragraph{Relevance (Score: 0 or 1):}
\begin{itemize}
    \item Are the highlighted image segments meaningfully related to the highlighted text components?
    \item Assess if the visual elements and the textual tokens work together to convey the intended message, especially in the context of the meme's overall meaning.
\end{itemize}

Please present your evaluation in the following format:

\textbf{Correctness:} [Score]\\
\textit{Justification:} [Your brief explanation]

\textbf{Relevance:} [Score]\\
\textit{Justification:} [Your brief explanation]
\end{tcolorbox}
\caption{Evaluation prompt provided to both human annotators and GPT-4V.}
\label{llm_evl_prompt}
\end{figure*}

\subsection{Inter-Annotator Agreement}\label{sec:intra-score}

We evaluated inter-annotator agreement among the three annotators by calculating Fleiss’ Kappa for both correctness and relevance. Table~\ref{tab:fleiss-kappa} presents the resulting values along with their interpretations.

\begin{table}[ht]
\centering
\resizebox{0.48\textwidth}{!}{%
\begin{tabular}{lcc}
\toprule
\textbf{Metric}    & \textbf{Fleiss’ Kappa Score} & \textbf{Interpretation} \\ 
\midrule
Correctness        & 0.7572                     & Substantial Agreement   \\ 
Relevance          & 0.6122                     & Moderate Agreement      \\ 
\bottomrule
\end{tabular}}
\caption{Fleiss’ Kappa scores for correctness and relevance.}
\label{tab:fleiss-kappa}
\end{table}

These findings indicate that the annotators achieved substantial agreement on correctness and moderate agreement on relevance. Overall, the results underscore the reliability of our annotation process.

\end{document}